\documentclass{article}

\usepackage[final]{neurips_2020}

\usepackage[utf8]{inputenc} % allow utf-8 input
\usepackage[T1]{fontenc}    % use 8-bit T1 fonts
\usepackage{hyperref}       % hyperlinks
\usepackage{url}            % simple URL typesetting
\usepackage{booktabs}       % professional-quality tables
\usepackage{amsfonts}       % blackboard math symbols
\usepackage{amsmath}
\usepackage{nicefrac}       % compact symbols for 1/2, etc.
\usepackage{microtype}      % microtypography
\usepackage{xcolor}
\usepackage{graphicx}
\usepackage{wrapfig}

\title{Making Graph Neural Networks Worth It for Low-Data Molecular Machine Learning}

\author{%
  Aneesh Pappu \\
  University College London\\
  \texttt{aneesh.pappu.19@ucl.ac.uk} \\
   \And
   Brooks Paige \\
   University College London \\
   Alan Turing Institute \\
   \texttt{b.paige@ucl.ac.uk} \\
}

\begin{document}

\maketitle

\begin{abstract}
  %{\color{red} TODO check all appendix references}
  Graph neural networks have become very popular for machine learning on molecules due to the expressive power of their learnt representations.
  However, molecular machine learning is a classically low-data regime and it isn’t clear that graph neural networks can avoid overfitting in low-resource settings. In contrast, fingerprint methods are the traditional standard for low-data environments due to their reduced number of parameters and manually engineered features. 
  In this work, we investigate whether graph neural networks are competitive in small data settings compared to the parametrically `cheaper’ alternative of fingerprint methods. 
  When we find that they are not, we explore pretraining and the meta-learning method MAML (and variants FO-MAML and ANIL) for improving graph neural network performance by transfer learning from related tasks.
  We find that MAML and FO-MAML do enable the graph neural network to outperform models based on fingerprints, providing a path to using graph neural networks even in settings with severely restricted data availability.
  In contrast to previous work, we find ANIL performs worse that other meta-learning approaches in this molecule setting. Our results suggest two reasons: molecular machine learning tasks may require significant task-specific adaptation, and distribution shifts in test tasks relative to train tasks may contribute to worse ANIL performance.
\end{abstract}

\section{Introduction}
\label{intro}
Quantitative structure-activity relationship (QSAR) modelling consists of fitting a molecular prediction model to predict biochemically relevant function. QSAR modelling has a rich history in the field of cheminformatics: experimentally screening candidate molecules can prove expensive, and computational methods play a significant role in screening novel candidates faster and cheaper by enabling prioritization of which compounds are worth experimentally testing.

% with computational methods serving to reduce cost and time during multiple stages of the drug development process
% In particular, medicinal chemists are often interested in optimizing therapeutic properties of candidate compounds without sacrificing other properties, which requires exploration of analog compounds. 
% {\color{red} I feel like most cheminformatics papers have this preamble of why in silico methods for drug discovery are useful. How much of this is necessary?}

Traditional approaches to QSAR prediction have relied on `fingerprint methods', which construct manually engineered bit vectors to represent molecular substructures. These representations are fed into off-the-shelf machine learning algorithms for downstream classification. The most popular standard in QSAR modelling is the ECFP/Morgan fingerprint \citep{Rogers2010, morgan_fingerprint_original_paper}, which we use in this work. However, since the Merck Molecular Activity Challenge, deep learning based approaches have become mainstream; in recent years, message passing neural networks, a variant of graph neural networks, have become a popular deep learning method for learning expressive representations of molecules, which are inherently graph structured \citep{Gilmer2017, chemprop_yang_swanson, kearnes2016molecular}. Message passing neural networks learn feature representations of nodes and/or edges from an input graph, which are often combined into a representation for the entire graph. A notable example which we use in this work is the Chemprop model of \citet{chemprop_yang_swanson}, which recently achieved state of the art results on a variety of molecular machine learning benchmarks. For details on Chemprop's message passing scheme, see Appendix A.

A main drawback of deep learning is that it requires large datasets to prevent overfitting, which is an issue in molecular machine learning because labeled biological data is expensive to collect and often sparse. Thus, manually engineered fingerprints combined with low parameter machine learning models have been traditionally well-suited for operating in low-data regimes. In particular, it has been shown that in low-data settings, fingerprint methods can outperform deep learning methods \citep{Mayr2018a, chemprop_yang_swanson}, so the issue of dataset size is of crucial importance when building machine learning models for molecular property prediction. As such, we first investigate whether deep learning methods can compete with traditional fingerprint alternatives in low-resource settings.

\begin{wrapfigure}{r}{0.45\textwidth}
\vspace{-2.5em}
\centerline{\includegraphics[width=0.95\linewidth]{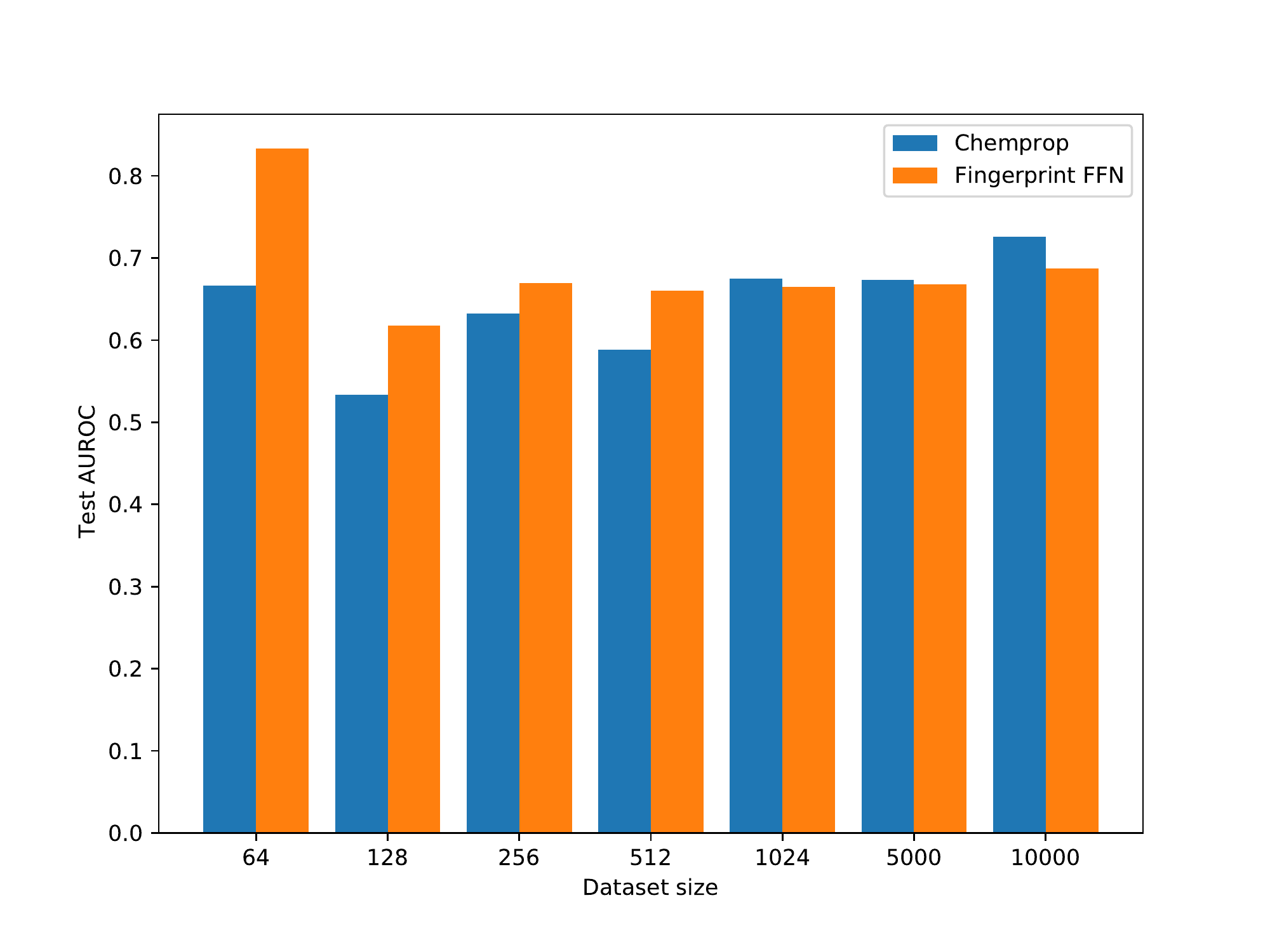}}
\caption{Performance of Chemprop versus the fingerprint method at varying dataset thresholds of the ChEMBL dataset. %{\color{red} Should this go in the appendix?}
}
\vspace{-0.5em}
\label{fig:chemprop_vs_ecfp}
\end{wrapfigure}

\section{Graph Neural Networks in Low Data Regimes}
To examine whether graph neural networks outperform fingerprint alternatives in low-data settings, we examine Chemprop's performance relative to an ECFP4 fingerprint architecture on the preprocessed ChEMBL20 dataset \citep{bento2014chembl, Mayr2018a} at varying dataset thresholds.
In both models, features computed by Chemprop and ECFP4 are inputs into a feed-forward network which outputs a property prediction.

Specifically, we compute the average AUROC of the fingerprint architecture and Chemprop trained jointly on all tasks with fewer datapoints than each threshold. These results are shown in Figure \ref{fig:chemprop_vs_ecfp}. We find that Chemprop is indeed outperformed by the fingerprint baseline at lower data thresholds, and as dataset size increases past 1024 datapoints, Chemprop becomes competitive and outperforms the fingerprint baseline. 
As the fingerprint method outperforms the graph neural network in smaller datasets, we shift our investigation into whether we can boost the performance of Chemprop in smaller data settings.

\section{Improving Low Data Performance of Graph Neural Networks}

\textbf{Pretraining} One strategy for mitigating deep learning's large data requirement is to leverage multitask pretraining, which combines data from multiple tasks to learn robust, general-purpose representations for future tasks. Perhaps the most famous example of pretraining is the practice of pretraining AlexNet \citep{krizhevsky2012imagenetalexnet} on the ImageNet classification dataset \citep{imagenetpaper}. Multitask learning has been explored in drug discovery and has seen success, with the caveat that models can exhibit positive or negative transfer on new tasks \citep{ramsundar2015massively}. We use the Chemprop architecture proposed in \citet{chemprop_yang_swanson} as our pretraining architecture (full hyperparameter details are in Appendix D). 

\textbf{Meta-Learning} Meanwhile, meta-learning methods \citep{schmidhuber1987evolutionary, hinton1987_metalearning_deblur, Vinyals2016_miniImageNet, lake2015human_omniglot, Altae-Tran2017a} have gained significant traction within the few-shot learning community, with the aim of leveraging prior tasks to `learn to learn' so that good performance on future tasks can be obtained with few datapoints. In particular, gradient-based meta-learning algorithms \citep{Ravi2017, Finn2017} have gained popularity, with perhaps the most recent significant advance coming from the Model Agnostic Meta-Learning (MAML) algorithm and its close variant First-Order MAML (FO-MAML) \citep{Finn2017}.

\textbf{MAML and FO-MAML} MAML and FO-MAML frame meta-learning as optimizing for a network initialisation that is capable of learning quickly on future tasks. MAML consists of a meta-training phase, where train tasks $T^{tr}$ are used to learn a meta-initialisation via an `outer loop' and `inner loop', and a meta-testing phase, where test tasks $T^{test}$ are used to evaluate how well the model adapts to new tasks. The inner loop calculates task specific updates to the meta-initialisation, and the outer loop calculates meta-gradients with respect to the meta-initialisation via the updated model parameters in order to update the meta-initialisation to an initialisation capable of learning quickly (further detail in Appendix B). As the outer loop update is quite expensive due to computing second-order gradients, \citet{Finn2017} propose FO-MAML, which omits second order gradient terms from the meta-update.

\textbf{ANIL: Feature Reuse vs. Rapid Learning} \citet{anil_raghu} conclude that MAML's efficacy arises from learning reusable features as opposed to features capable of rapid learning. They found that across supervised and reinforcement learning tasks, MAML-trained initialisations adapt very little during test task-specific learning, as evidenced by high CCA similarity between pre and post adaptation layers. As a result, \citet{anil_raghu} propose the ANIL algorithm, which removes the inner loop for all layers but the head layer, and observe near identical performance to MAML.

\textbf{Dataset and Experiment} We pretrain Chemprop and implement MAML, FO-MAML, and ANIL on Chemprop to see whether these approaches boost Chemprop's performance relative to the fingerprint method. We filter the ChEMBL20 \citep{bento2014chembl, Mayr2018a} dataset by tasks that have between 128 and 1024 datapoints. This results in a new dataset consisting of 645 binary classification tasks across 5 distinct task types obtained from the ChEMBL database: ADME (A), Toxicity (T), Binding (B), Functional (F), and Unassigned (U). Following \citet{gsk_maml_paper}, we split the 645 tasks into three task splits, $T^{tr}, T^{val}, T^{test}$. $T^{test}$ is composed of 10 randomly assigned B and F tasks and all of the A, T, and U tasks. $T^{val}$ is randomly assigned 10 B and F tasks, and the remaining B and F tasks are assigned to $T^{tr}$. This allows us to assess performance of our methods on both in-distribution tasks (by performance on the B and F tasks in $T^{test}$) and out-of-distribution tasks (by performance on the A, T, and U tasks in $T^{test}$). We use a scaffold split for all \textit{within} task splits. A summary of the task split is shown in Table \ref{tab:meta_task_split_numbers} and further dataset details are included in Appendix C.

\textbf{Evaluation} The performances of the fingerprint method, pretrained Chemprop, and all meta-learning methods are shown in Table \ref{tab:chembl_in_dist_perf} and Table \ref{tab:chembl_out_of_dist_perf} for the in-distribution and out-of-distribution tasks, respectively. We summarize the average rank of each method in Table \ref{tab:average_ranks} and use Wilcoxon signed-rank testing to assess significance of rank differences on all pairwise method performances, with p-values shown in Table \ref{tab:wilcoxon_table}. As shown by average ranks and Wilcoxon testing, we find that pretraining \textit{doesn't} significantly outperform the fingerprint baseline. However, we find that MAML and FO-MAML \textit{significantly outperform} both the pretraining and fingerprint methods, showing that meta-learning is able to boost the performance of the graph neural network in this low-data setting. 
% {\color{red} Took out comment on positive and negative transfer -- not our focus}
%  The average ranks suggest that pretraining shows positive transfer on in-distribution test tasks and negative transfer on the out-of-distribution test tasks relative to the fingerprint method, but across all tasks the performance is not significantly different as shown in Table \ref{tab:wilcoxon_table}.

Interestingly, we find that ANIL performs
consistently worse than MAML and is not statistically significantly different from the fingerprint or pretrained models (Table \ref{tab:wilcoxon_table}). This is in contrast to the findings of \citet{anil_raghu} that ANIL suffers no loss in performance compared to MAML. This discrepancy motivates our next investigation into why ANIL performs poorly in this molecule setting.

\begin{table}[h]
\small
\caption{In distribution mean AUPRC with standard deviations across five folds. Best performing result is bold text and second best is regular text. `Frac. Pos.' denotes the fraction of positives in the dataset (expected AUPRC of a random classifier). `\# Obs' is number of datapoints in the task.}
\resizebox{\textwidth}{!}{%
\begin{tabular}{cccccccc}
\hline
ChEMBL ID & ECFP                   & Pretraining            & MAML                   & FO-MAML                & ANIL                   & Frac. Pos. & \# Obs \\ \hline
1738202   & {\color{gray}0.953 $ \pm $ 0.086}          & {\color{gray}0.909 $ \pm $ 0.050}      & \textbf{0.997 $ \pm $ 0.006} & 0.991 $ \pm $ 0.013          & {\color{gray}0.942 $ \pm $ 0.032}          & 0.965             & 144           \\
3215176   & 0.542 $ \pm $ 0.270  & {\color{gray}0.362 $ \pm $ 0.097}    & {\color{gray}0.482 $ \pm $ 0.342}          & \textbf{0.576 $ \pm $ 0.271} & {\color{gray}0.431 $ \pm $ 0.360}          & 0.083             & 157           \\
1963934   & \textbf{0.957 $ \pm $ 0.022} & {\color{gray}0.865 $ \pm $ 0.091}          & {\color{gray}0.863 $ \pm $ 0.113}          & {\color{gray}0.883 $ \pm $ 0.057 }         & 0.928 $ \pm $ 0.026          & 0.964             & 165           \\
1794358   & {\color{gray}0.264 $ \pm $ 0.218 }       & {\color{gray}0.415 $ \pm $ 0.331}          & \textbf{0.577 $ \pm $ 0.295} & 0.517 $ \pm $ 0.321          & {\color{gray}0.337 $ \pm $ 0.336  }        & 0.059             & 222           \\
2114797   & \textbf{0.868 $ \pm $ 0.107} & {\color{gray}0.805 $ \pm $ 0.073  }        & {\color{gray}0.796 $ \pm $ 0.036 }         & 0.827 $ \pm $ 0.074          & {\color{gray}0.569 $ \pm $ 0.041}          & 0.576             & 224           \\
3215116   & {\color{gray}0.167 $ \pm $ 0.070}          & {\color{gray}0.152 $ \pm $ 0.078 }         & {\color{gray}0.210 $ \pm $ 0.117  }        & 0.292 $ \pm $ 0.218          & \textbf{0.484 $ \pm $ 0.179} & 0.161             & 248           \\
1794355   & 0.935 $ \pm $ 0.068          & \textbf{0.954 $ \pm $ 0.027} & {\color{gray}0.932 $ \pm $ 0.029     }     & {\color{gray}0.895 $ \pm $ 0.045   }       & {\color{gray}0.932 $ \pm $ 0.035  }        & 0.947             & 304           \\
1614202   & {\color{gray}0.897 $ \pm $ 0.086 }         & 0.928 $ \pm $ 0.084          & {\color{gray}0.900 $ \pm $ 0.076    }      & {\color{gray}0.874 $ \pm $ 0.115 }         & \textbf{0.957 $ \pm $ 0.047} & 0.927             & 314           \\
1794567   & {\color{gray}0.920 $ \pm $ 0.036  }        & \textbf{0.939 $ \pm $ 0.051} & {\color{gray}0.923 $ \pm $ 0.067   }       & {\color{gray}0.923$ \pm $ 0.068 }          & 0.924 $ \pm $ 0.078          & 0.904             & 385           \\
1614359   & \textbf{0.728 $ \pm $ 0.147} & {\color{gray}0.621 $ \pm $ 0.117   }       & 0.702 $ \pm $ 0.167          & {\color{gray}0.681 $ \pm $ 0.168   }       & {\color{gray}0.628 $ \pm $ 0.185   }       & 0.497             & 390           \\
1738131   & {\color{gray}0.476 $ \pm $ 0.085     }     & {\color{gray}0.446 $ \pm $ 0.121 }         & 0.556 $ \pm $ 0.146          & \textbf{0.631 $ \pm $ 0.087} &{\color{gray} 0.444 $ \pm $ 0.171   }       & 0.314             & 468           \\
1614170   & {\color{gray}0.364 $ \pm $ 0.102  }        & {\color{gray}0.360 $ \pm $ 0.118  }        & \textbf{0.456 $ \pm $ 0.088} & 0.395 $ \pm $ 0.155          & {\color{gray}0.310 $ \pm $ 0.093  }        & 0.311             & 546           \\
1963705   & {\color{gray}0.562 $ \pm $ 0.109     }     & {\color{gray}0.775 $ \pm $ 0.045  }        & 0.807 $ \pm $ 0.075          & {\color{gray}0.765 $ \pm $ 0.078  }        & \textbf{0.862 $ \pm $ 0.046} & 0.441             & 692           \\
1909212   & {\color{gray}0.049 $ \pm $ 0.021  }        & 0.150 $ \pm $ 0.155          & \textbf{0.185 $ \pm $ 0.121} & {\color{gray}0.078 $ \pm $ 0.034 }         & {\color{gray}0.050 $ \pm $ 0.017 }         & 0.017             & 824           \\
1909209   & {\color{gray}0.264 $ \pm $ 0.136  }        & {\color{gray}0.155 $ \pm $ 0.088  }        & 0.527 $ \pm $ 0.192          & \textbf{0.544 $ \pm $ 0.276} &{\color{gray} 0.275 $ \pm $ 0.164 }         & 0.077             & 835           \\
1909085   & {\color{gray}0.247 $ \pm $ 0.155  }        & {\color{gray}0.408 $ \pm $ 0.176 }         & \textbf{0.811 $ \pm $ 0.169} & 0.668 $ \pm $ 0.267          &{\color{gray} 0.254 $ \pm $ 0.118 }         & 0.079             & 835           \\
1909192   & {\color{gray}0.013 $ \pm $ 0.004  }        & \textbf{0.278 $ \pm $ 0.370} & 0.231 $ \pm $ 0.385          & {\color{gray}0.045 $ \pm $ 0.019 }         & {\color{gray}0.026 $ \pm $ 0.037 }         & 0.004             & 838           \\
1909092   & {\color{gray}0.066 $ \pm $ 0.073}          &{\color{gray} 0.084 $ \pm $ 0.099  }        & 0.580 $ \pm $ 0.366          & \textbf{0.797 $ \pm $ 0.284} & {\color{gray}0.059 $ \pm $ 0.077 }         & 0.029             & 838           \\
1909211   & {\color{gray}0.497 $ \pm $ 0.134 }         & 0.780 $ \pm $ 0.072          & \textbf{0.802 $ \pm $ 0.108} & {\color{gray}0.771 $ \pm $ 0.133 }         & {\color{gray}0.423 $ \pm $ 0.134 }         & 0.112             & 839           \\
1963741   & {\color{gray}0.461 $ \pm $ 0.053 }         & {\color{gray}0.554 $ \pm $ 0.141   }       & 0.682 $ \pm $ 0.075          & \textbf{0.718 $ \pm $ 0.080} & {\color{gray}0.644 $ \pm $ 0.060 }         & 0.330             & 919 \\ \hline
\end{tabular}}
\label{tab:chembl_in_dist_perf}
\end{table}

\begin{table}[h]
\small
\caption{Out of distribution mean AUPRC with standard deviations across five folds. Best performing result is bold text and second best is regular text. `Frac. Pos.' denotes the fraction of positives in the dataset (expected AUPRC of a random classifier). `\# Obs' is number of datapoints in the task.}
\resizebox{\textwidth}{!}{%
\begin{tabular}{cccccccc}
\hline
ChEMBL ID & ECFP          & Pretraining   & MAML          & FO-MAML       & ANIL          & Frac. Pos. & \# Obs \\ \hline
2098499   & 0.516$ \pm $0.274 & {\color{gray}0.500$ \pm $0.159} & {\color{gray}0.508$ \pm $0.190} & \textbf{0.593$ \pm $0.188} & {\color{gray}0.483$ \pm $0.231} & 0.255             & 137           \\ 
1738021   & {\color{gray}0.899$ \pm $0.079} & 0.958$ \pm $0.033 & \textbf{0.973$ \pm $0.025} & {\color{gray}0.888$ \pm $0.088} & {\color{gray}0.885$ \pm $0.118} & 0.891             & 138           \\ 
1738019   & {\color{gray}0.731$ \pm $0.153} & {\color{gray}0.682$ \pm $0.136} & {\color{gray}0.791$ \pm $0.088} & \textbf{0.903$ \pm $0.127} & 0.793$ \pm $0.158 & 0.752             & 165           \\ 
918058    & {\color{gray}0.282$ \pm $0.361} & {\color{gray}0.205$ \pm $0.196} & 0.429$ \pm $0.393 & \textbf{0.626$ \pm $0.458} & {\color{gray}0.036$ \pm $0.016} & 0.067             & 225           \\ 
2095143   & 0.440$ \pm $0.309 & {\color{gray}0.074$ \pm $0.062} & {\color{gray}0.418$ \pm $0.309} &\textbf{ 0.539$ \pm $0.282} & {\color{gray}0.418$ \pm $0.290} & 0.062             & 273           \\ 
2028077   & {\color{gray}0.040$ \pm $0.011} & {\color{gray}0.057$ \pm $0.023} & {\color{gray}0.300$ \pm $0.254} & 0.494$ \pm $0.353 & \textbf{0.508$ \pm $0.324} & 0.038             & 289           \\ \hline
\end{tabular}}
\label{tab:chembl_out_of_dist_perf}
\end{table}

\begin{table}[h]
\centering
\caption{Average rank of each method within the subsets of in-distribution tasks, out-of-distribution tasks, and all tasks. Bold is best (lowest) average rank and regular text is second best average rank.}
\begin{tabular}{cccccc}
\hline
Task Subset & ECFP & Pretraining & MAML & FO-MAML  & ANIL \\ \hline
In-distribution     & {\color{gray}3.55}   & {\color{gray}3.25} & \textbf{2.20} & 2.50  & {\color{gray}3.50}     
\\
Out-of-distribution & {\color{gray}3.17 }                    & {\color{gray}4.00 }          & 2.67   & \textbf{1.67 }& {\color{gray}3.50 }     
\\
All         & {\color{gray}2.70}           & {\color{gray}2.10}         & \textbf{1.43}   & 1.79          & {\color{gray}2.37}                     \\ \hline
\end{tabular}
\label{tab:average_ranks}
\end{table}

\section{Feature Reuse vs. Rapid Learning in Molecular Machine Learning}

\textbf{CCA Similarity} To investigate ANIL's poor performance, we study task adaptation versus feature reuse via measuring layer representation similarity pre and post adaptation as in \citet{anil_raghu}. Specifically, we use our MAML trained initialisation to adapt the network on each test task across 5 seeds, and calculate the CCA similarity between each layer of the graph neural network pre and post adaptation. As the message passing network maintains a variable number of hidden representations based on the number of atoms in a molecule, we calculate similarity based on the final molecule level representation of the graph neural network, which is size invariant to the number of atoms in the molecule. We also compute CCA similarity coefficients for both feed-forward layers.

\begin{wrapfigure}{r}{0.58\textwidth}
\vspace{-45pt}
\centerline{\includegraphics[width=0.9\linewidth]{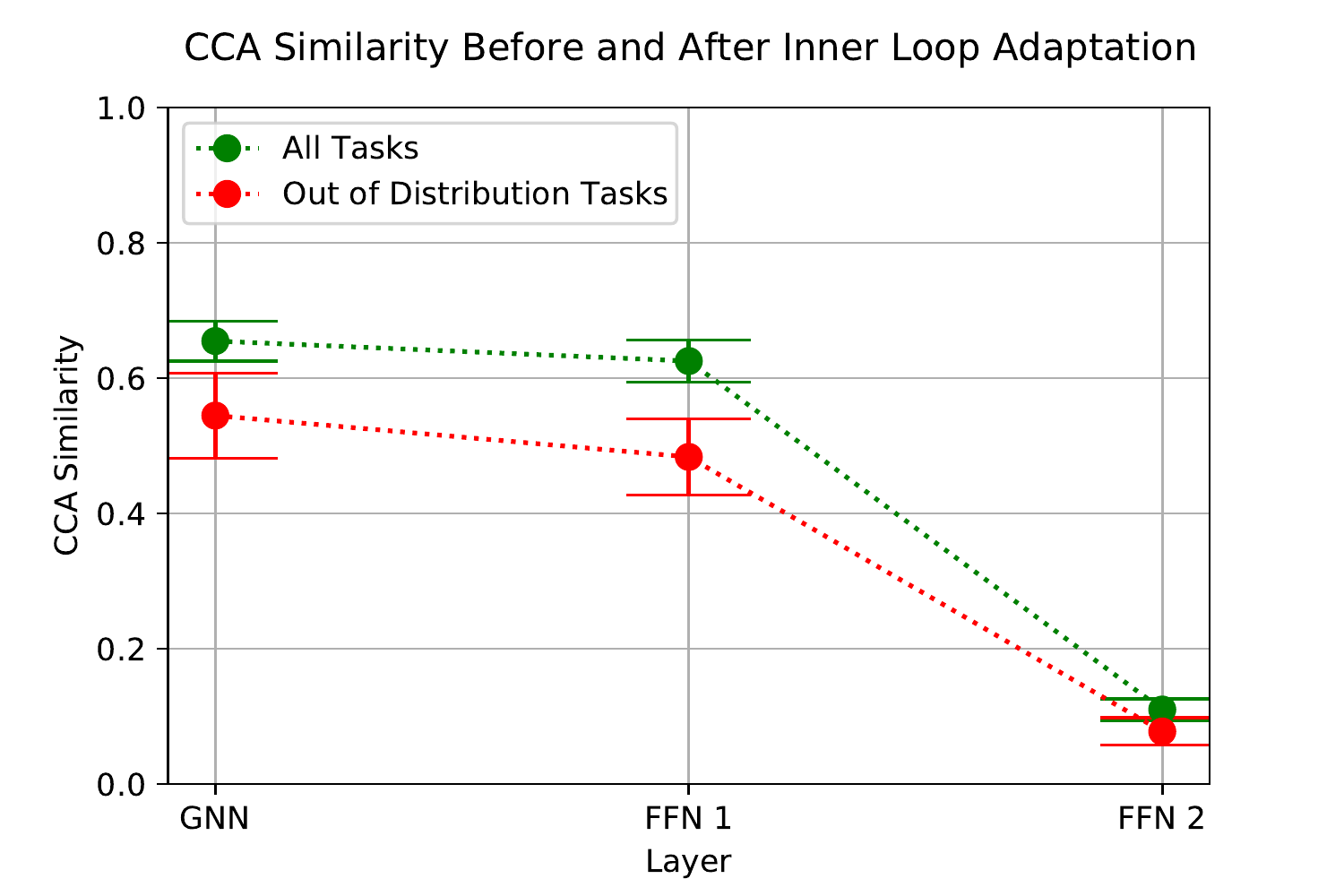}}
\caption{CCA similarity pre and post inner loop adaptation across all tasks and solely the out of distribution tasks.}
\label{fig:cca_similarity}
\end{wrapfigure}

The similarity results are shown in Figure \ref{fig:cca_similarity}. Compared to the CCA similarity experiment in \citet{anil_raghu}, the average similarity for each layer is much lower than what we would expect if inner loop adaptation wasn't necessary. \citet{anil_raghu} observes that for all layers except the head, CCA similarity is nearly 1. In contrast, we see that the graph neural network and first feed-forward layer have changed significantly after inner loop adaptation. These findings suggest that ANIL performs poorly in this molecule setting because inner loop adaptation is necessary for the body layers, and that distribution shift relative to the meta-train tasks adversely affects ANIL's effectiveness, as there is decreased layer similarity when solely examining the out of distribution tasks.

\section{Conclusion and Future Work}
\label{conclusion}
In this work we investigated whether graph neural networks are useful compared to fingerprint methods in low-data settings. We found that while graph neural networks underperform relative to fingerprint methods, MAML and FO-MAML significantly improve performance, outperforming both the fingerprint and pretraining methods. This shows meta-learning enables use of graph neural networks in low-data settings over fingerprint methods. We also find that ANIL, contrary to prior belief, does not match performance of MAML. Our results suggest that inner loop adaptation is important in this molecule setting and that ANIL's performance is affected by distribution shift in test tasks. Future work includes investigating whether ANIL breaks in the supervised and reinforcement learning settings investigated in \citet{anil_raghu} when domain shift is engineered into the meta-task splits.

\newpage

\begin{ack}
We would like to acknowledge Cuong Nguyen and the GSK Artificial Intelligence team, Kyle Swanson, Bharath Ramsundar, Yihong Chen, Luca Franceschi, and Pasquale Minervini for helpful comments and feedback. This work was supported by The Alan Turing Institute under the EPSRC grant EP/N510129/1. We would also like to thank the Marshall Aid Commemoration Commission for providing a Marshall Scholarship to fund AP.
\end{ack}

%%%%%%%%%%%%%%%%%%%%%%%%%% references section

\small
\newpage

\bibliographystyle{apalike}
\bibliography{references}

%%%%%%%%%%%%%%%%%%%%%%%%%% references section

\newpage
\normalsize
\section*{Appendix}

\section*{A \hspace{5mm} Chemprop Message Passing Details}
\label{chemprop_message_passing}

Concretely, for an input graph G with nodes $v, w$, message update function $M_t$ and hidden state update function $U_t$, each iteration of message passing computes updates as
\begin{align}
m_{vw}^{t+1} &= \sum_{k \in \{ N(v) \setminus w \}} h_{vw}^t,
&
h_{vw}^{t+1} &= \tau(h_{vw}^0 + W_m m_{vw}^{t+1})
\end{align}

where $\tau$ is the ReLU nonlinearity \citep{relu_paper}, and $W_m$ is a learnable matrix. The initial hidden state is computed as $h_{vw}^0 = \tau(W_i(x_v, e_{vw})) $ where $W_i$ is a learnable matrix and $x_v$ and $e_{vw}$ correspond to input node and edge features. After the prespecified iterations of message passing have concluded, Chemprop computes the atom-level representations as 
\begin{align}
m_v &= \sum_{k \in N(v)} h_{kv}, &
h_v &= \tau(W_a (x_v, m_v))
\end{align}
where $W_a$ is a learnable matrix. Finally, the graph-level representation is calculated via a sum-pool operation, $h_G = \sum_{v \in G} h_v$. Afterwards, the readout phase computes the prediction $\hat y$ by applying a feed-forward neural network to the graph representation.
\section*{B \hspace{5mm} MAML Algorithm Details}
During meta-training, the model is initialised with parameters $\theta$ and a batch of tasks $\{T_i\}$ is sampled. For each task, a \textit{support} and \textit{query} set are sampled. The model is updated with respect to the loss on the support set of the parameters $\theta$ using the standard SGD rule:
\[
    \theta_i^{'} = \theta - \alpha \nabla_{\theta}L_{T_i}(\theta) \]
where $\theta_i^{'}$ are the new parameters for task $T_i$ computed using the gradient of the loss on the support set of task $T_i$ and $\alpha$ is the inner loop learning rate. This is repeated for each meta-train task, producing a set of parameters $\{\theta_i^{'}\}$ for each meta-train task $T_i$. These updated parameters $\theta_i^{'}$ are then used to calculate the loss on the query set for each task which represents a task-specific validation loss. This loss acts as the meta-loss, which is used to calculate the gradient with respect to the \textit{original} parameters $\theta$ to update the meta-initialisation $\theta$:
\[    \theta = \theta - \beta \nabla_\theta \sum_{T_i}L_{T_i}(\theta_i^{'})\]
where $\beta$ is the outer loop learning rate.

\section*{C \hspace{5mm} Meta Task Splits and Dataset Details}
\label{meta_task_sp}
Table \ref{tab:meta_task_split_numbers} shows the number of tasks in each task type allocated to each meta-task split. 

\begin{table}[h]
\caption{Number of tasks in each meta-task split by task type.}
\label{tab:meta_task_split_numbers}
\center
\begin{tabular}{llllll}
\hline
                & \textit{A} & \textit{T} & \textit{U} & \textit{B} & \textit{F}   \\ \hline
$T^{train}$   & 0 & 0 & 0 & 128 & 471 \\ \hline
$T^{val}$ & 0 & 0 & 0 & 10  & 10  \\ \hline
$T^{test}$       & 2 & 2 & 2 & 10  & 10  \\ \hline
\end{tabular}
\end{table}

In the meta-train and meta-validation splits, we use a within-task split ratio of 80\% train and 20\% validation. In the meta-test split we use a split ratio of 80\% train, 10\% validation, and 10\% test. This is because during meta-training and meta-validation, we only need a support set for task-specific adaptation and a query set for calculating meta-loss and meta-gradients, but in the meta-test phase we require a train and validation set for training and early stopping, and a test set for calculating final test performance. For the pretraining method, we combine $T^{tr}$ and $T^{val}$ which is then split into $D^{tr}$ for training and $D^{val}$ for early stopping and hyperparameter tuning. We train the fingerprint architecture from scratch for each of the test tasks in $T^{test}$. Each method is evaluated over 5 seeds on each task in $T^{test}$.

\section*{D \hspace{5mm} Hyperparameters}
\label{appC}
We use the ADAM optimizer \citep{kingma2014adam} for all optimizations. We use Learn2Learn \citep{arnold2019learn2learn} and PyTorch \citep{pytorch_ref} for our meta-learning implementations.
\subsection*{D.1 \hspace{5mm} Fingerprint Architecture} We use ECFP4 fingerprints for our input fingerprints. The final fingerprint hyperparameters after grid search are a dropout rate of 0.2 and a hidden layer size of 400. We use a batch size of 32 and a learning rate of $10^{-4}$ for all fingerprint experiments. 

\subsection*{D.2 \hspace{5mm} Pretraining Architecture}
The final pretraining hyperparameters after grid search are a dropout rate of 0.2 and 2 message passing steps. We fix the number of feed-forward layers used in the readout phase of the graph neural network to be 2 so that it matches the fingerprint baseline architecture. We use a batch size of 32 and a learning rate of $10^{-4}$ for all experiments.

\subsection*{D.3 \hspace{5mm} Meta-Learning Models}
The meta-learning models use the same hyperparameters as the pretraining architecture as the meta-learning methods are implemented on top of Chemprop. The primary meta-learning hyperparameters we tune via grid search are the outer and inner loop learning rates. In this work we use an outer loop learning rate of $10^{-3}$ and an inner loop learning rate of $0.05$. At meta-test time, we use a learning rate of $10^{-4}$. We use a meta batch size of 32 (i.e., each outer loop update happens using meta-gradients on 32 tasks), and an inner loop batch size of 32 (i.e., 32 datapoints per task).

\section*{E \hspace{5mm} Wilcoxon Signed Rank Test}
\label{appD}

Table \ref{tab:wilcoxon_table} shows the p-values computed for each pairwise grouping of methods. In particular, each method is represented as a sample of 130 datapoints where each datapoint is the AUPRC on one seed on one of the 26 test tasks. We represent each method by this vector $\in \mathbb{R}^{130}$ for the Wilcoxon signed-rank test. We use a significance threshold of $0.05$.

\begin{table}[h!]
\centering
\caption{P-values of Wilcoxon signed-rank test performed pairwise across all methods. Signed-rank test computed with respect to each method's performance across all 26 test tasks and the 5 seeds on each task. Threshold for significance is 0.05, and significant p-values are bold.}
\begin{tabular}{l|lllll}
\hline
            & ECFP           & Pretraining     & MAML            & FO-MAML        & ANIL           \\ \hline
ECFP        &           *    & 0.73            & \textbf{0.002}  & \textbf{0.001} & 0.83           \\ 
Pretraining & 0.73           &    *             & \textbf{0.0003} & \textbf{0.004} & 0.99           \\
MAML        & \textbf{0.002} & \textbf{0.0003} &             *    & 0.47           & \textbf{0.01}  \\ 
FO-MAML     & \textbf{0.001} & \textbf{0.004}  & 0.47            &         *       & \textbf{0.004} \\ 
ANIL        & 0.83           & 0.99            &  \textbf{0.01}   & \textbf{0.004} &         *      \\ 
\hline
\end{tabular}
\label{tab:wilcoxon_table}
\end{table}

\end{document}